\documentclass[runningheads]{llncs}
\usepackage{graphicx}
\usepackage{amsmath}
\usepackage{booktabs}
\usepackage[ruled,vlined]{algorithm2e}
\begin{document}
\title{Imperatives for Virtual Humans}
\author{Weizi Li and Jan M. Allbeck}
\authorrunning{Li and Allbeck}
\institute{Laboratory for Games and Intelligent Animation \\
George Mason University \\
4400 University Drive, MSN 4A5 \\
Fairfax, VA 22030 \\
\email{\{wlia,jallbeck\}@gmu.edu}}
\maketitle              % typeset the header of the contribution
\begin{abstract}
Seemingly since the inception of virtual humans, there has been an effort to make their behaviors more natural and human-like. In additions to improving movement's visual quality, there has been considerable research focused on creating more intelligent virtual characters. This paper presents a framework inspired by natural language constructs that aims to author more reasonable virtual human behaviors using structured English input. We focus mainly on object types and properties, quantifiers, determiners, and spatial relations. The framework provides a natural, flexible authoring system for simulating human behaviors.

\end{abstract}

\section{Introduction}
Since the inception of virtual humans, there has been an effort to make their behaviors more natural, realistic, and human-like. Some research has focused on improving the visual quality of animated movements~\cite{Li2012Distribution}, such as walking, reaching, gesturing, and facial expressions. Other work has focused on connecting motions together in meaningful ways, resulting in longer-duration behaviors for denoting more autonomous and intelligent virtual characters~\cite{Li2011Purpose}. One can think of a continuum between pure animation research and artificial intelligence research as applied to virtual humans. Certainly, there has been a fair amount of work focusing on making more intelligent virtual characters~\cite{Li2012Commonsense}. For example, in the foundational work, Funge et al. created characters that could reason and plan their behaviors~\cite{funge1999cognitive}; Kallmann and Thalmann worked on creating more intelligent behaviors of agents by making their interactions with objects smarter~\cite{kallmann1999direct}; Badler et al. researched on creating smart avatars using natural language instructions as the input~\cite{badler1999animation}.

The work presented in this paper is inspired by all of the previous efforts to make virtual characters more realistic and natural, but in particular, we are motivated to use natural language constructs to create and animate the behaviors of virtual humans. Effective instructions include a lot of vital information (some explicitly, some implicitly). To start with, there are the core semantics of instructed actions (e.g., go somewhere, pick something up, operate something). There may also be a structure of sub-actions (e.g., hold the nut while
loosening the bolt). %The participants, agents and objects, are also important. Implicitly or explicitly, initiation and termination conditions indicate what should be true in the virtual world to start and stop the instructed action(s) [7]. 
In this paper, we mainly focus on the \emph{context} of an instruction: What context does the instruction include and how does it match the virtual environment? Consider, for example, \emph{Pickup the briefcase in the corner of the office}. Using the article \emph{the} implies that there will be a single object that fits the other criteria specified, namely an object that is a briefcase being located in the corner of an office. Furthermore, as humans, we learn and accept that \emph{corner} may not have a strict, calculated definition. If we attempt to execute this command and find a briefcase near a corner of the office, we are likely to accepted a fuzzy definition of \emph{corner} and continue execution. If, however, the briefcase is more in the middle of the office, we may abort executing the instruction and ask for clarification.

In this paper, we present a framework for interactively instructing virtual humans using structured English. Users can input instructions through the GUI or structured imperatives in a text file. A virtual human will then attempt to follow the instructions, comparing the context outlined in the instructions to what they encounter in the virtual world and raising warnings when there are ambiguities and mismatches. Similar to the work of Bindiganavale et al.~\cite{bindiganavale2000dynamically}, our framework includes semantic specifications for actions and objects with links to motion generators and graphical models. Our instructional system utilizes object types, stored in a hierarchy, as well as object properties such as color. After discussing some related work, we will present how our framework processes spatial prepositions, including spatial regions and object relative locations. We will then present the affects of quantifiers and determiners. Finally, we will outline a few examples before concluding and discussing future work.

\section{Related Work}
There exists substantial work in studying and incorporating the usage of spatial prepositions. The first set of work is in the field of linguistics.~\cite{bates1976spatial,hershkovits1986language,talmy1983language} provide early studies in spatial prepositions in English. As these work's authors and other researchers have noted, modeling spatial relations and especially using natural language to capture them is extremely complex. Even so, for decades, many sophisticated computational models have been proposed and developed. Andr\'{e} et al.~\cite{andre1986characterizing,andre1986coping} developed the German dialog system CITYTOUR and focused on describing spatial relations between city constructions. Kelleher et al.~\cite{kelleher2009applying} built two computational models of topological and projective spatial prepositions to visually situate dialog. Virtual Guide, another dialog system built by Hofs et al.~\cite{hofs2010natural} includes an embodied conversational agent who can assist users in finding their way in a virtual environment. For interested readers, a review of more applications exploiting spatial prepositions can be found in~\cite{baldwin2009prepositions}.

Natural Language Processing, a research field closely related to linguistics, has also been exploring spatial relationships. For example, an extension of a linguistic ontology has been used in~\cite{bateman2010linguistic}. Particularly, the authors provide detailed semantics aiming to generate a Generalized Upper Model in simplifying spatial relationship computing. In~\cite{kordjamshidi2011spatial}, researchers use a \emph{spatial role labeling} technique, providing the machine the ability to extract spatial prepositions, trajectories, and landmarks, allowing it to learn spatial relations via a specific learning method. Djalali et al.~\cite{djalali2012corpus} built an online game, Cards Corpus, in a 2D grid-based environment, where two players have to navigate each other in order to solve a cards collection task. Players can only communicate via chat windows. Their speech transcripts are captured and can be used for studying real human usage of spatial prepositions.

Computing and interpreting spatial relations is also important for robotics research, especially for navigation and path-finding. Kress-Gazit et al.~\cite{kress2010automatic} have integrated a Structured English interface, including several locative prepositions, into a Linear Temporal Logic system for synthesizing controllers to guide robots. In~\cite{tellex2011understanding}, though authors are not specifically addressing spatial prepositions, they have adopted a probabilistic graphical model for specific natural language commands for robotic navigation in semi-structured environments. In addition, Hawes et al.~\cite{hawes2012towards} built a cognitive system for recognizing context-dependent spatial regions in real-world scenarios. These spatial regions include \emph{Front/Back Rows}, \emph{Kitchen}, and \emph{Office} with the assistance of characteristic physical objects.

In animation research, Xu et al.~\cite{xu2000algorithms} designed computational models for several directional spatial prepositions in order to generate motion trajectories for virtual agents. In~\cite{bindiganavale2000dynamically}, authors introduce mechanisms for real humans using natural language to alter behaviors of their avatars in a virtual environment. There also exist work in using text as input to generate a virtual scenario. CarSim~\cite{dupuy2001generating} is a system producing 3D simulations of car accidents from written descriptions. Coyne et al.~\cite{coyne2001wordseye} developed a system for converting text into static 3D scenes.

In this paper, we are not focusing on natural language processing. Instead, we focus on operational aspects of resolving spatial prepositions and other language constructs to disambiguate and animate imperatives for virtual humans.

\section{Approach}
\begin{figure}
\centering
\includegraphics[width=\textwidth]{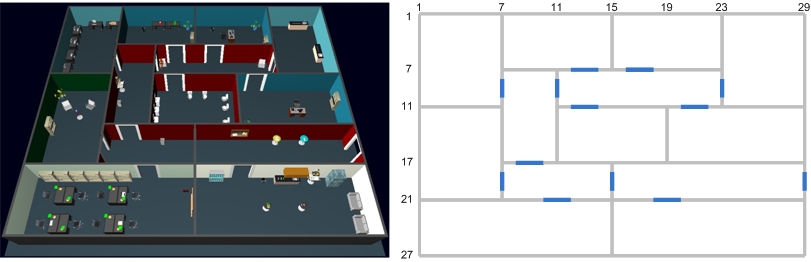}
\caption{An overview of our virtual environment and its corresponding 2D layout.} 
\label{fig:layout}
\end{figure}

In this section, we first detail our implementation of spatial relationships with both locations and objects, then introduce a realization of quantifiers and determiners. The virtual environment we are exploring and its corresponding 2D layout are shown in Figure~\ref{fig:layout}.

\subsection{Spatial Relationships with Locations}
For each location in our environment, we compute four spatial regions: \emph{Corner}, \emph{End}, \emph{Middle}, and \emph{Side}. Given a conceptually, grid-based environment, we have
$G_{width}$ and $G_{length}$ indicating how many grids will be take into account for any of the four spatial regions. Their values depend on location border’s width $B_{width}$ and length $B_{length}$. In addition, users can alter $G_{width}$ and $G_{length}$ ($0<G_{width}\le \frac{B_{width}}{2},0<G_{length}\le \frac{B_{length}}{2}$) to meet their specific needs. Default values for $G_{width}$ and $G_{length}$ are as follows:

\begin{equation*}
G_{width(length)} = 
\begin{cases}
  1, \text{if } 0<B_{width(length)}\le 4\\      
  2, \text{if } 4<B_{width(length)}\le 8\\   
  3, \text{if } 8<B_{width(length)}\le 12\\   
  4, \text{if } 12<B_{width(length)}\le 16\\  
  ...
\end{cases}
\end{equation*}

\begin{figure}
\centering
\includegraphics[width=\textwidth]{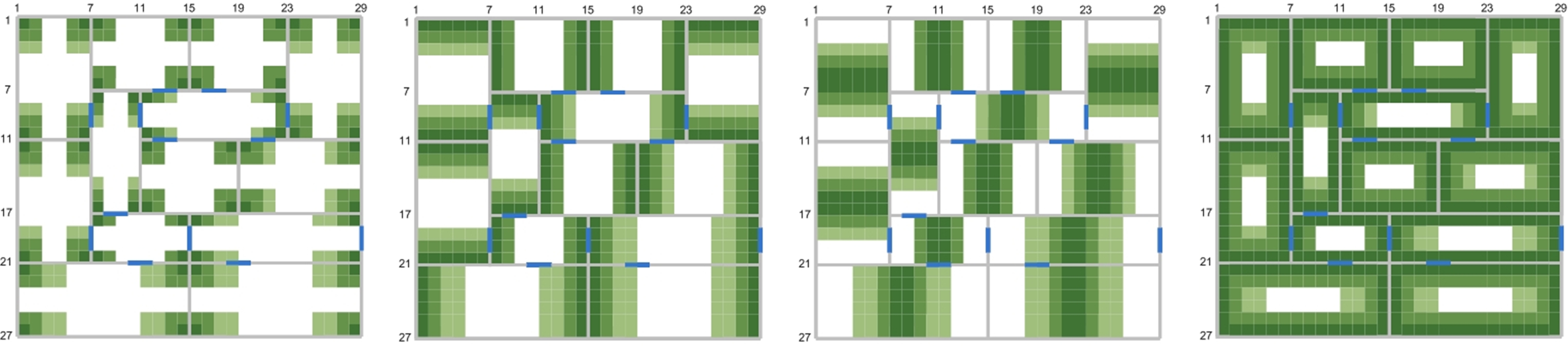}
\caption{Spatial relationships with locations: from left to right are results of \emph{Corner}, \emph{End}, \emph{Middle}, and \emph{Side}.} 
\label{fig:spatial-location}
\end{figure}

To provide an illustration, for \emph{Corner}, if the current room's width is 6 and length is 10, we have $G_{width} = 2$ and $G_{length} = 3$. Then, for each corner in this room we have $G_{width} \times G_{length} = 6$ grid nodes. The algorithm for computing the number of grid nodes for a corner is shown in Algorithm~\ref{alg:spatial-location}. The computation for the rest of the spatial regions (i.e., \emph{End}, \emph{Middle}, and \emph{Side}) is similar. The overall result is visually demonstrated in Figure~\ref{fig:spatial-location}.

\begin{algorithm}[H]
\SetAlgoLined
 // G[n]: all grid nodes in the current location \\
 // Corner[\,]: grid nodes in a corner, initially empty \\
 // Loc: the current location \\
 \For{$i$ such that $0 \le i < n$}{
  \If{$Loc.startZ < G[i].z < Loc.startZ + G_{length}$ \textbf{\upshape{and}}
      $Loc.endZ - G_{length} < G[i].z < Loc.endZ$}{
   \If{$Loc.startX < G[i].x < Loc.startX + G_{width}$ \textbf{\upshape{and}} 
      $Loc.endX - G_{width} < G[i].x < Loc.endX$}{
    $Corner[\,] \leftarrow G[i]$
   }}{}
 }
 \caption{Calculate the number of grid nodes for \emph{Corner}}
 \label{alg:spatial-location}
\end{algorithm}

As one may notice, in each spatial region, we have three different colors of grid nodes. These three colors from the darkest to the brightest represent three fuzzy quantifiers: \emph{Strict}, \emph{Proximate} and \emph{Near}, respectively. To be specific, the grid node with the darkest color in the corner indicates that it is the strictest definition of a corner (i.e., \emph{Strict}). We implemented this feature, because we observe that human spatial perceptions are fuzzy~\cite{freeman1975modelling} and also natural language is often ambiguous. Many factors involving the characteristics of locations and observers themselves could influence the interpretation of spatial relations. In addition, people tend to use vague constraints, or in other terms, the fuzzy quantifiers, rather than precise numbers in describing spatial regions~\cite{lakoff2008women}, for instance, ``Stand in the corner'' or ``The table approximately in the middle of this room''. Encoding degrees of these spatial relationships enables both the input instructions and the resulting animated behaviors to be more human-like. Instructions can specify a degree to which the relation holds or they can be under-specified, in which case the system will attempt to resolve them \emph{strictly}, \emph{proximately}, or \emph{nearly}. For example, if an agent is told to ``Stand in the corner'' and all of Strict corner nodes are blocked, it will then try a Proximate corner node and, following that, a Near corner node. Additionally, we have implemented \emph{Left}, \emph{Right}, \emph{Near}, and \emph{Far} to facilitate an even more robust instructional framework. With these terms, the agent can now perform commands such as ``Go to the far right corner'' or ``Go to the left side''. The calculation of these spatial relations is based on the agent's position and orientation when entering a room.

\subsection{Spatial Relationships with Respect to Objects}
\begin{figure}
\centering
\includegraphics[width=.8\textwidth]{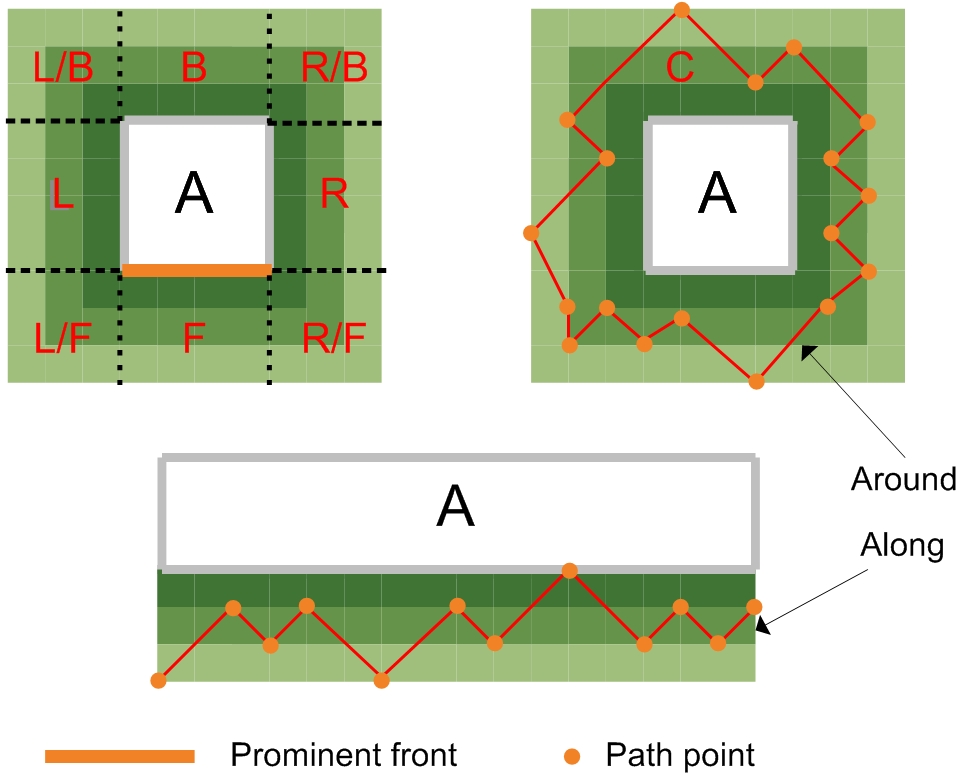}
\caption{Spatial relationships with objects: In front of, Behind, To the left of, To the right of, Close to, Along and Around (L--left, R--right, F--front, B--back, C--close
to; Above and Under are not showing).} 
\label{fig:spatial-object}
\end{figure}

For objects, we compute the following spatial relationships: \emph{In front of}, \emph{Behind}, \emph{To the left of}, \emph{To the right of}, \emph{Above}, \emph{Under}, and \emph{Close to}. In addition, we define two spatial paths \emph{Along} and \emph{Around}. The first four spatial relations usually depend on either the prominent front of (the bounding box of) the object (deictic use of the relationship) or speaker's current position (intrinsic use of the relationship)~\cite{andre1986coping}. Here, we only consider the deictic use of the relations since we believe if an object has a easy-determine front, when giving navigation instructions, most people will take the prominent front as the reference rather than using their current position. These relations, with \emph{Above} and \emph{Under}, which we compare the center of bounding box of the involving objects, are all under the relationship \emph{Close to}. In addition, the dynamic paths \emph{Along} and \emph{Around} are defined by a linkage of randomly selected path points also under the relationship \emph{Close to}. An overall illustration is presented in Figure~\ref{fig:spatial-object}.

Last but not least, given that goals are differ, for both locations and objects, we are not intend to develop a comprehensive quantitative and computational model for recognition of spatial relationships, which is a research focus of many researchers. Instead, we aim to design a system such that the users can easily change various parameters to fit different scenarios and satisfy their own definitions of different spatial relationships.

\subsection{Quantifiers}
Quantifiers can add richness to an agent's behavior. Our framework includes several quantifiers: all, a lot of, many, several, a few, a couple and any. A mapping to specific values is shown in Table~\ref{tb:q}. The precise values can easily be edited. Once a quantifier has been selected, the framework attempts to identify a sufficient number of instances of objects meeting the overall specification. If there is an insufficient number of objects, the agent will simply act on all of the objects available and the system will notify the user of the shortfall.

\begin{table*}[ht] %(h = here, t = top)
    \centering
    \scalebox{1}{
    \begin{tabular}{ccccc}
        \toprule  
        \textbf{Quantifier} & \textbf{Value}  \\
         \midrule
        all & maximum \\
         \midrule
        a lot of & 10 \\
         \midrule
        many & 8 \\
         \midrule
        several & 6 \\
         \midrule
        a few & 4 \\
         \midrule
        a couple & 2 \\
         \midrule
        any & 1 \\
        \bottomrule
    \end{tabular}}
    \caption{Mapping between quantifiers and specific values.}
    \label{tb:q}
\end{table*}

\subsection{Determiners}
In addition to quantifiers, our system allows instructions to include a number of determiners: 
\begin{itemize}
    \item \textbf{a/an}: either \textbf{a} or \textbf{an} will return one qualified object instance.
    \item \textbf{the}: Returns the object instance that meets the specifications. If there are no other specifications or more than one object meets them, one of the qualifying objects is selected and the user is notified of the discrepancy.
    \item \textbf{the only}: Similar to \textbf{the}, a stronger warning is presented to the user.
    \item \textbf{the same}: A history of object interactions is used to disambiguate this determiner. If no previous objects meet the specification of the object, a warning message is generated.
    \item \textbf{different}: This determiner also uses the history of object interactions. It attempts to find an object that satisfies the specification and has not been used for the same action by the same agent previously. If none can be found, a warning message is generated.
    \item \textbf{both}: If only one object instance exists, return it; If exact two objects exist, return both of them; If there are more than two objects, then return two of them. Warning messages are generated if there are not precisely two objects that meet the specification.
    \item \textbf{either}: If only one object instance exists, return it; If two or more objects exist, return one of them. Warning messages are generated if there are not precisely two objects that meet the specification.
    \item \textbf{your}: While determiner quantification is rare~\cite{roeper2006acquisition}, in our system, this is the only determiner that can be used in combination with quantifiers. This operation will return all qualified belongings of selected agent.
\end{itemize}

\section{Examples}
\begin{figure}
\centering
\includegraphics[width=\textwidth]{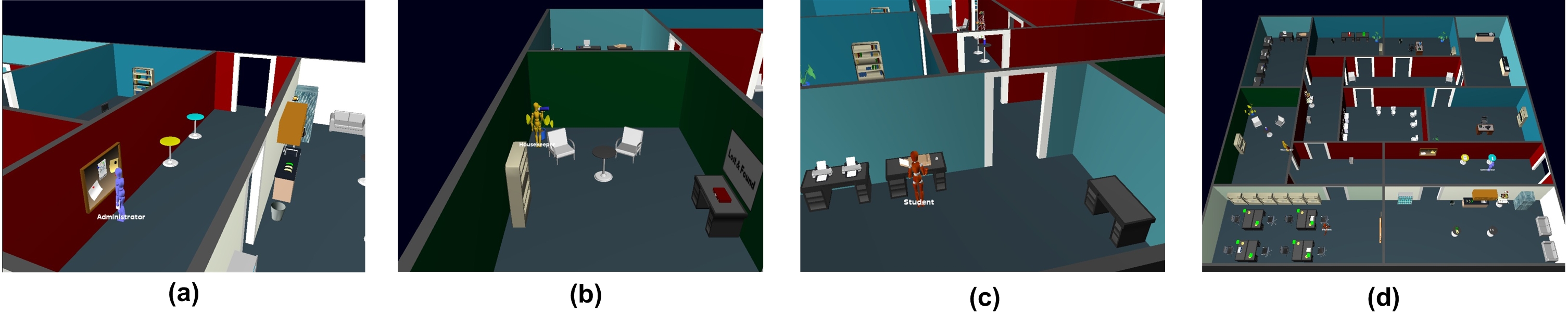}
\caption{Single agent scenario (operating with GUI): (a) ``Eat a couple of yellow bananas.''; (b) ``Eat a few green bananas above the round table.''; (c,d) ``Pickup all blue
mice that are near a monitor and keyboard in the strict far right corner of Laboratory 0.''.} 
\label{fig:ex1}
\end{figure}

\begin{figure}
\centering
\includegraphics[width=\textwidth]{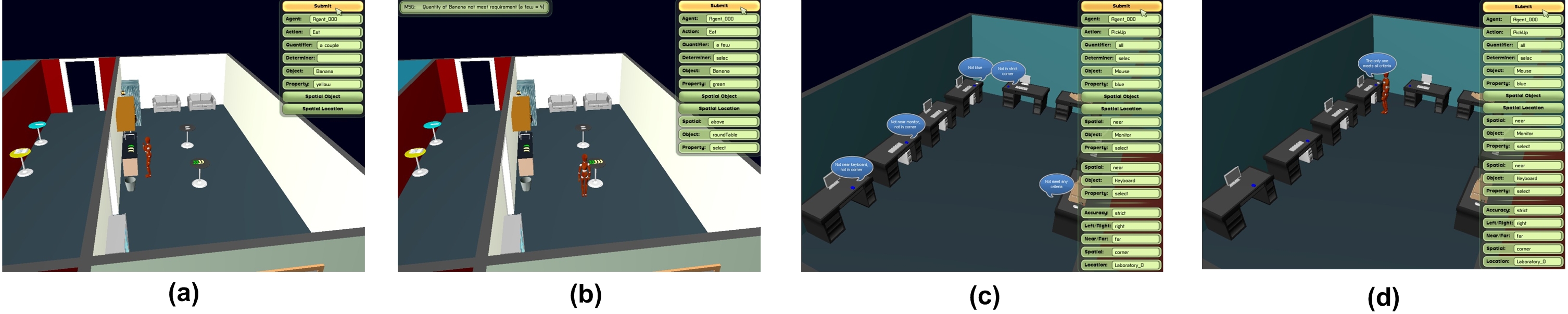}
\caption{Multiple agents scenario (operating with text files as input): (a) the administrator is pinning up a poster; (b) the housekeeper is watering a yellow plant; (c) the
student is filling paper for the copy machine near a mouse; (d) a bird's eye view of the scene.} 
\label{fig:ex2}
\end{figure}

In this section, we will provide several examples demonstrating our framework. The first set of examples are instructions for a single agent delivered through the GUI. As we can see in Figure~\ref{fig:ex1}(a), using the GUI, an agent is instructed to eat a couple of yellow bananas. Here, since the user does not specify any spatial relationships and all requirements have been meet (i.e., there exists yellow bananas in the scene and their quantity exceeds a couple, according to Table~\ref{tb:q},
the agent will perform the specified behavior and no warning message will be generated. In contrast, in Figure~\ref{fig:ex1}(b), the agent is asked to eat a few green bananas above a round table. Although there are some green bananas in the environment, their quantity fails to meet a few which has value 4. In response, there is a warning message generated in the top left corner of the screen notifying the user. In addition, since there is a spatial relationship specified (i.e., above a round table), the agent will not eat green banana above the square desk, but only those above the round table. A more complex example is shown in Figure~\ref{fig:ex1}(c,d), where multiple spatial relationships have been selected in reference to both a room and objects. The agent is able to determine the object fitting all of the requirements. 

The second set of example instructions include multiple agents using text files as the input. Sample screenshots are shown in Figure~\ref{fig:ex2}. Here, we have three agents with roles of Administrator, Housekeeper, and Student, respectively. Each one of them acts based on instructions provided by the text file. For simplicity, here, we only show the content of the Administrator text file: ``1) Carry the only mail above the round cyan table in near middle of Hallway 1. 2) Pick up the only poster above the round yellow table in strict middle of Hallway 1. 3) Pin up the poster on billboard in Hallway 1. 4) Deliver the mail to the green container on the far side of Office 0.''

\section{Conclusion and Future Work}
We have presented a framework for instructing virtual humans through a natural language inspired interface. We have focused on object types and properties, quantifiers, determiners, and spatial relations. The framework provides a natural, flexible authoring system for simulating human behaviors. Users can interactively instruct and modify agent behaviors through a structured English GUI or by providing the same text in a file. Such a file could be thought of as a script (similar to a theatrical script) for a cast of agents. While performing agents will adhere to the script as closely as possible, but ask for a clarification if the script does not match the current environment, just as real humans might.

Future work includes loosening the structured requirement of the English input by integrating available natural language parsers, semantic taggers, and pragmatic processing units. Though not a solved problem, we believe natural language processing tools, along with environmental disambiguation will help clarifying prepositional phrase attachment issues, which we currently handle only sequentially. Also, we plan to add more spatial relations and test their results against user expectations in a variety of contexts. Finally, we are interested in integrating this work with agents who have enhanced memory~\cite{Li2013Memory} and learning capabilities~\cite{Li2012Apprentice}.

\bibliographystyle{ieeetr}
\bibliography{ref}

\end{document}